\documentclass[lettersize,journal]{IEEEtran}

\usepackage{cite}
\usepackage{pifont}
\newcommand{\xmark}{\ding{55}}
\usepackage{amsmath,amssymb,amsfonts}
\usepackage{algorithmic}
\usepackage{graphicx}
\usepackage{booktabs} 

\let\labelindent\relax

\usepackage{enumitem}
\usepackage{tabularx}
\usepackage{xcolor}
\usepackage{soul}
\usepackage{multirow}
\usepackage{graphicx}
\usepackage{gensymb}
\usepackage{textcomp}
\DeclareRobustCommand{\hlred}[1]{{\sethlcolor{pink}\hl{#1}}}

\DeclareRobustCommand{\hlgreen}[1]{{\sethlcolor{green}\hl{#1}}}
\usepackage{comment}
\usepackage{algorithmic}
\usepackage{ifpdf}
\usepackage{url}
\usepackage{caption}
\usepackage{subcaption}

\begin{document}

\twocolumn[{%
\vspace{20mm}
{ \large
\begin{itemize}[leftmargin=2.5cm, align=parleft, labelsep=2cm, itemsep=4ex,]

\item[\textbf{Citation}]{K. Kokilepersaud, S. Trejo Corona, M. Prabhushankar, G. AlRegib, C. Wykoff,  "Clinically Labeled Contrastive Learning for OCT Biomarker Classification," in \textit{IEEE Journal of Biomedical and Health Informatics,} 2023.}

\item[\textbf{Review}]{Date of Publication: May 18th 2023}

\item[\textbf{Codes}]{\url{https://github.com/olivesgatech/OLIVES_Dataset.git}}

\item[\textbf{Bib}]  {@inproceedings\{kokilepersaud2023clinically,\\
    title=\{Clinically Labeled Contrastive Learning for OCT Biomarker Classification\},\\
    author=\{K. Kokilepersaud, S. Trejo Corona, M. Prabhushankar, G. AlRegib, C. Wykoff\},\\
    booktitle=\{IEEE Journal of Biomedical and Health Informatics\},\\
    year=\{2023\}\}}


\item[\textbf{Contact}]{
\{kpk6, mohit.p, alregib\}@gatech.edu\\\url{https://ghassanalregib.info/}\\}
\end{itemize}

}}]
\newpage

\title{Clinically Labeled Contrastive Learning for OCT Biomarker Classification}
\author{Kiran Kokilepersaud, \IEEEmembership{Student Member, IEEE},  Stephanie Trejo Corona, 
        Mohit Prabhushankar, \IEEEmembership{Member, IEEE}, 
        Ghassan AlRegib, \IEEEmembership{Fellow, IEEE}, and Charles Wykoff
\thanks{This work was submitted on  June 23, 2022.}
\thanks{Kiran Kokilepersaud, Mohit Prabhushankar, and Ghassan AlRegib are with the Omni Lab for Intel. Visual Eng. \& Science (OLIVES) at the Center for Signal \& Info. Processing (CSIP) at the Georgia Institute of Technology, Atlanta, GA 30308 USA  (e-mail: \{kpk6, mohit.p, alregib\}@gatech.edu) }
\thanks{Stephanie Trejo Corona and Charles Wykoff are with the Retina Consultants of Texas, Houston, TX 77339 USA (e-mail:\{stephanie.trejo, ccwmd\}@retinaconsultantstexas.com)}}

\maketitle

\begin{abstract}

This paper presents a novel positive and negative set selection strategy for contrastive learning of medical images based on labels that can be extracted from \emph{clinical data}. In the medical field, there exists a variety of labels for data that serve different purposes at different stages of a diagnostic and treatment process. Clinical labels and biomarker labels are two examples. In general, clinical labels are easier to obtain in larger quantities because they are regularly collected during routine clinical care, while biomarker labels require expert analysis and interpretation to obtain. Within the field of ophthalmology, previous work has shown that clinical values exhibit correlations with biomarker structures that manifest within optical coherence tomography (OCT) scans. We exploit  this relationship by using the clinical data as pseudo-labels for our data without biomarker labels in order to choose positive and negative instances for training a backbone network with a supervised contrastive loss. In this way, a backbone network learns a representation space that aligns with the clinical data distribution available. Afterwards, we fine-tune the network trained in this manner with the smaller amount of biomarker labeled data with a cross-entropy loss in order to classify these key indicators of disease directly from OCT scans. We also expand on this concept by proposing a method that uses a linear combination of clinical contrastive losses. We benchmark our methods against state of the art self-supervised methods in a novel setting with biomarkers of varying granularity. We show performance improvements by as much as 5\% in total biomarker detection AUROC. 

\end{abstract}


\section{Introduction}
\label{sec:intro}
\begin{figure}[t]
\centering
\includegraphics[scale=.3]{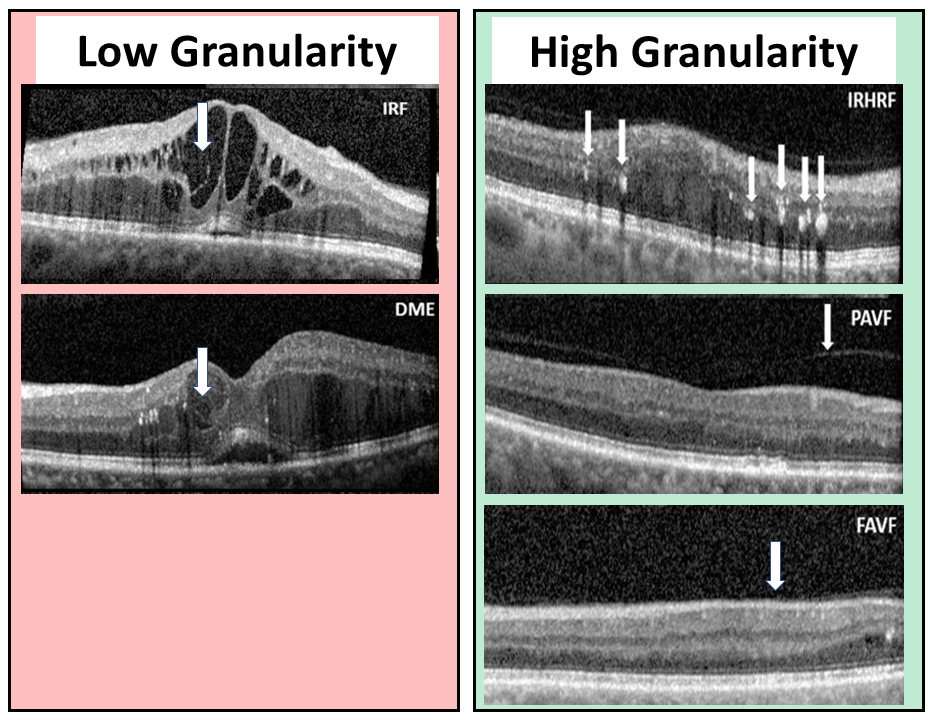}

\caption{This is a demonstration of the difference in granularity levels between different biomarkers. Biomarkers such as IRF and DME exhibit a low granularity, meaning it stands out clearly with respect to the rest of the image. Biomarkers such as FAVF, PAVF, and IRHRF exist as small localized structures that are typically harder to detect due to their higher granularity. An in-depth discussion of biomarkers in OCT can be found at \cite{markan2020novel}.\vspace{-.3cm}}

\label{fig: examples}
\end{figure}

\IEEEPARstart{C}{ontrastive} learning \cite{le2020contrastive} refers to a family of self-supervised algorithms that leverages differences and similarities between data points in order to extract useful representations for downstream tasks. The basic premise is to train a model to produce a lower dimensional space where similar pairs of images (positives) project much closer to each other than dissimilar pairs of images (negatives). Due to its nonexistent dependence on labels, contrastive learning has seen interest within the medical field \cite{xu2021review} where access to sufficient labels is scarce and expensive \cite{ghassemi2020review}. One potential application area for contrastive learning is within the context of biomarker detection for indicators of disease within Optical Coherence Tomography (OCT) scans. Biomarkers refer to “any substance, structure, or
process that can be measured in the body or its products and
influence or predict the incidence of outcome or disease \cite{strimbu2010biomarkers}.” The major bottleneck towards producing models that can aid physicians in finding and assessing biomarkers, such as those found in Figure \ref{fig: examples}, is the lack of access to a large labeled training pool. This is due to the requirement of a trained expert to perform labeling, thus motivating the potential application of contrastive representation learning on a larger unlabeled set before fine-tuning a model on lesser available labeled data.  However, contrastive learning, in its basic form, does not account for several practical considerations that surround the setting of medical data. The main aspects of this setting that remain unexplored are the presence of biomarkers of varying granularity and the presence of a wide variety of clinical information that exhibits relationships with these biomarkers.

\begin{figure}[t]
\centering
\includegraphics[scale=.4]{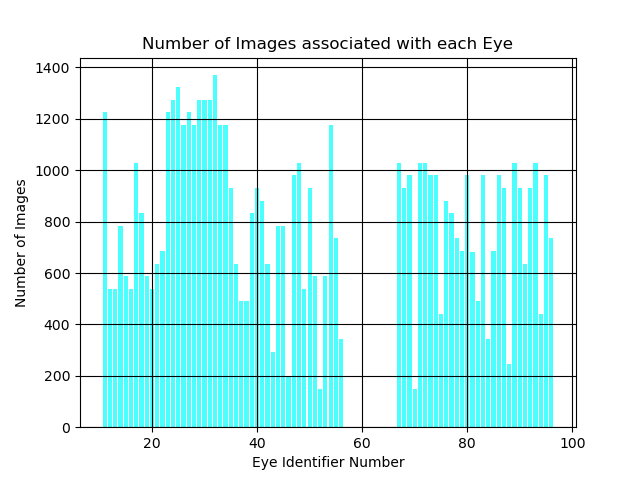}

\caption{Histogram of Eye/Patient image distribution within \texttt{OLIVES} \cite{prabhushankar2022olives} dataset.}

\label{fig:patient_dist}
\end{figure}

\begin{figure}[t]
\centering
\includegraphics[scale=.4]{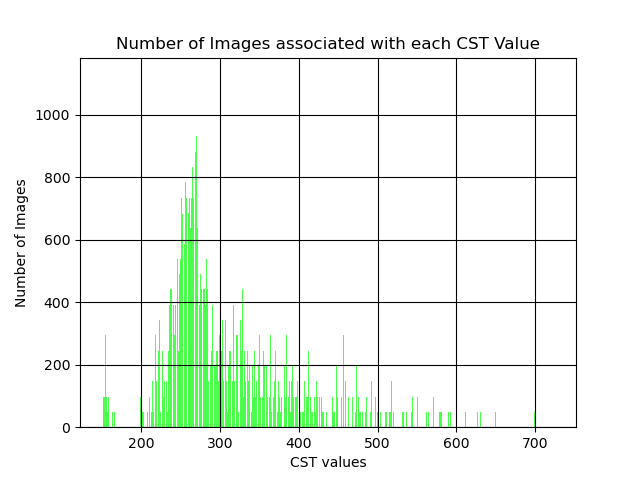}

\caption{Histogram of CST image distribution within \texttt{OLIVES} \cite{prabhushankar2022olives} dataset.}

\label{fig:bcva_dist}
\end{figure}

\begin{figure}[t]
\centering
\includegraphics[scale=.4]{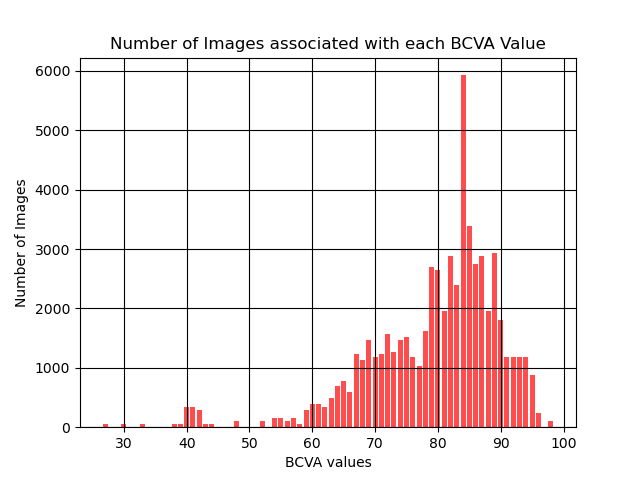}

\caption{Histogram of BCVA image distribution within \texttt{OLIVES} \cite{prabhushankar2022olives} dataset.}

\label{fig:cst_dist}
\end{figure}

\begin{figure}
     \centering
     \begin{subfigure}[b]{0.5\textwidth}
         \centering
         \includegraphics[scale = .25]{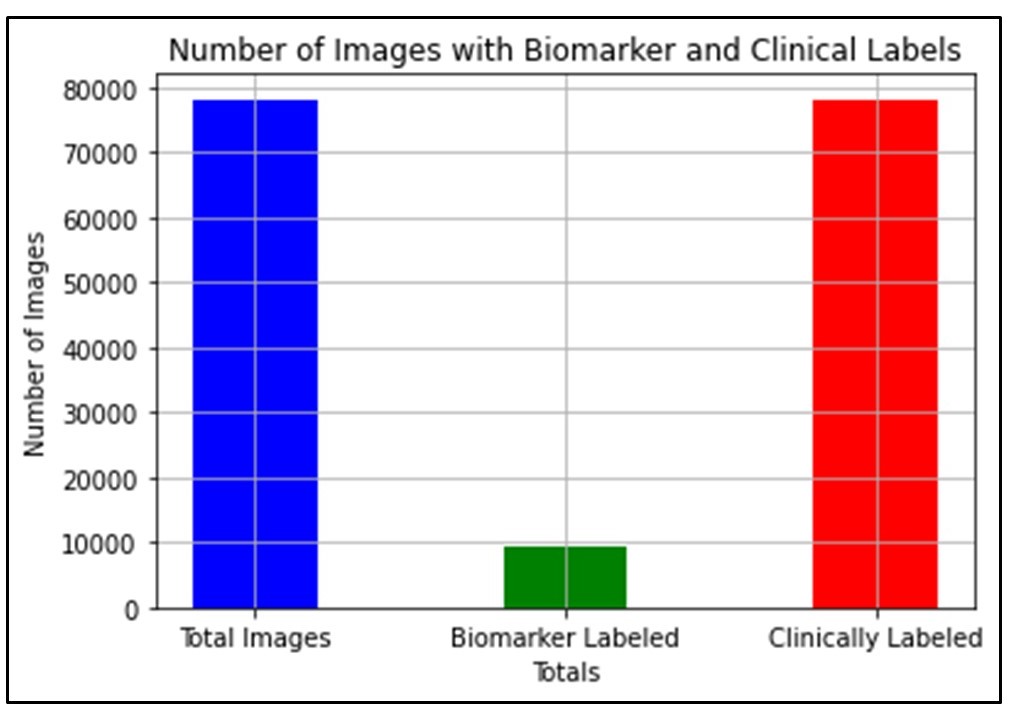}
         \caption{This shows
the number of images with biomarker and clinical labels in the OLIVES dataset.}
         \label{fig:quantity}
     \end{subfigure}
     \hfill
     \begin{subfigure}[b]{0.5\textwidth}
         \centering
         \includegraphics[scale=.4]{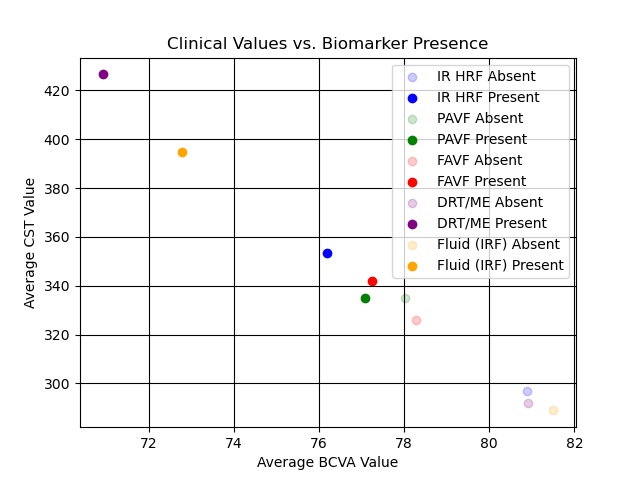}
         \caption{ All 9408 OCT
scans with biomarker labels were grouped based on the presence or absence of a specific biomarker.
These biomarker groups were then averaged based on their associated CST and BCVA values. It
can be observed that, on average, images with a biomarker present are separable from images with
a biomarker absent, with respect to clinical values, thus indicating a relationship between clinical
values and biomarkers.}
         \label{fig:seperable}
     \end{subfigure}
        \caption{This gives an overview of statistics regarding biomarkers and clinical labels.}
        \label{fig:three graphs}
\end{figure}

Previous work \cite{cole2021does} has shown that contrastive learning performs worse as the granularity of the task increases. This is a relevant setting in the medical domain as shown in Figure \ref{fig: examples}. It is observed that while certain biomarkers exhibit a low granularity where the biomarker of interest is clearly distinguishable from the rest of image, other biomarkers are small localized fine-grained structures that exhibit a high granularity. Ideally the algorithm applied should work regardless of granularity level, but no previous work has explored this granularity problem within an explicit medical context.

\begin{table*}[]
\centering
\begin{tabular}{@{}ccccc@{}}
\toprule
\multicolumn{1}{c}{Method} &
  \multicolumn{1}{c}{\begin{tabular}[c]{@{}c@{}}Patient\\ Pair Selection\end{tabular}} &
  \multicolumn{1}{c}{\begin{tabular}[c]{@{}c@{}}Clinical \\ Pair Selection\end{tabular}} &
  \multicolumn{1}{c}{\begin{tabular}[c]{@{}c@{}}Medical\\ Granularity Analysis\end{tabular}} &
  \multicolumn{1}{c}{\begin{tabular}[c]{@{}c@{}}Combined Clinical\\ Losses\end{tabular}} \\ \midrule
Traditional Contrastive Learning Methods \cite{chen2020simple,chen2020improved,li2020prototypical} & \xmark & \xmark & \xmark & \xmark \\ \midrule
Previous Patient Integration Attempts \cite{vu2021medaug,azizi2021big,diamant2021patient}                     & \checkmark & \xmark & \xmark & \xmark \\ \midrule

   Clinical Contrastive (Ours)          & \checkmark & \checkmark & \checkmark & \checkmark \\\bottomrule
   
\end{tabular}
\caption{Comparison of attempts at utilizing contrastive learning within medical domain.}
\label{tab:contrast_comparison}
\end{table*}
Another pitfall of current contrastive learning techniques is that previous approaches operate under the assumption that there exists simply labeled and
unlabeled data, rather than distributions of various types of labels. As a result, traditional approaches, such as \cite{chen2020simple}, rely on data augmentations to generate positive and negative pairs of data. However, in medical data, there oftentimes exists data that is naturally collected during routine clinical care that may act as a useful surrogate for selecting positive and negative pairs. To illustrate this point, we show statistics of available data from the \texttt{OLIVES} dataset for ophthalmology  \cite{prabhushankar2022olives} in Figure \ref{fig:quantity}. It can be observed that of the 78108 Optical Coherence Tomography (OCT) scans within this dataset, all are labeled with some type of clinical information collected from standard clinical practice, while a small amount is labeled with biomarker information that required explicit expert annotation. This motivates the question of whether this clinically labeled data can be integrated in some manner. Recent attempts at utilizing clinical information within contrastive learning frameworks \cite{vu2021medaug,azizi2021big,diamant2021patient} have tried to utilize the images associated with individual patients as a means to choose positives and negatives in the contrastive loss. While these approaches have seen success compared to traditional strategies, they do not consider other potential clinical label distributions that may better inform the positive pair selection process as well as the potential to use multiple types of clinical information in tandem with each other. This can be visualized by the histograms from the \texttt{OLIVES} dataset. Figure \ref{fig:patient_dist} shows the setting of previous attempts where positive pairs could potentially be chosen from images associated with the same patient or eye, but Figures \ref{fig:bcva_dist} and \ref{fig:cst_dist} show distributions from collected Best Central Visual Acuity (BCVA) and Central Subfield Thickness (CST) values that also have the potential for selecting good positive pairs in a contrastive loss. This premise is further  supported by Figure \ref{fig:seperable} that shows the existence of a relationship between the BCVA and CST values and whether each corresponding biomarker exists or not. Furthermore, medical studies such as \cite{hannouche2012correlation,sun2014disorganization,murakami2011association,kashani2010retinal} confirm that values collected during clinical procedures 
can act as indicators of structural changes that manifest
within associated imaging data. All of this motivates the potential that clinical information, beyond just the patient identity, has as a means for choosing positive and negative pairs within a contrastive loss.  A summary of these research gaps can be found in Table \ref{tab:contrast_comparison}. With these considerations in mind, the contributions of this paper are as follows:

\begin{enumerate}
  \item We introduce for the first time how ophthalmic clinical values, beyond just the patient identity, can be utilized for choosing positive and negative pairs in a contrastive loss.
  
  \item We introduce an approach that uses a combined contrastive loss on multiple clinical values and show that utilizing this combined loss is more robust to a variety of perturbations in the training and testing setup. 

  \item We compare and show performance improvements against state of the art contrastive learning approaches in a novel varying biomarker granularity setting.
  
  \item We analyze our approach in settings that include: varying data access, individual and multi-label biomarker detection, across splits of patients, semi-supervised, and across different encoder architectures.
  
\end{enumerate}

\label{sec:intro}

\section{Theoretical Interpretation}

In \cite{arora2019theoretical} the authors present a theoretical framework for contrastive learning. Let $X$ denote the set of all possible data points. In this framework, contrastive learning assumes access to similar data in the form of $(x,x^{+})$ that comes from a distribution $D_{sim}$ as well as $k$ iid negative samples $x^{-}_{1}, x^{-}_{2}, ..., x^{-}_{k}$ from a distribution $D_{neg}$. This idea of similarity is formalized through the introduction of a set of latent classes $C$ and an associated probability distribution $D_{c}$ over $X$ for every class $c\in C$. $D_{c}(x)$ quantifies how relevant $x$ is to class $c$ with a higher probability assigned to data points belonging to this class. Additionally, let us define $\rho$ as a distribution that describes how these classes naturally occur within the unlabeled data. From this, the positive and negative distribution are characterized as $D_{sim} = \displaystyle \mathop{\mathbb{E}}_{c \sim \rho} D_{c}(x)D_{c}(x^+)$ and $D_{neg} = \displaystyle \mathop{\mathbb{E}}_{c \sim \rho} D_{c}(x^-)$ where $D_{neg}$ is from the marginal of $D_{sim}$. 


The key idea that separates our work from the standard contrastive learning formulation presented above is a deeper look at the relationships between $\rho$, $D_{sim}$, and $D_{neg}$. In principal, during unsupervised training, there is no information that provides the true class distribution $\rho$ of the dataset $X$. The central goal of contrastive learning is to generate an effective $D_{sim}$ and $D_{neg}$ such that the model is guided towards learning $\rho$ by identifying the distinguishing features between the two distributions. Ideally, this guidance occurs through the set of positives belonging to the same class $c_{p}$ and all negatives belonging to any class $c_{n} \neq c_{p}$ as shown in the supervised framework \cite{khosla2020supervised}. Traditional approaches such as \cite{chen2020simple,chen2020improved,li2020prototypical}, enforces positive pair similarity through augmenting a sample to define a positive pair which would clearly represent an instance belonging to the same class. However, these strategies do not define a process by which negative samples are guaranteed to belong to different classes. This problem is discussed in \cite{arora2019theoretical} where the authors decompose the contrastive loss $L_{un}$ as a function of an instance of a hypothesis class $f \in F$ into $L_{un}(f) = (1-\tau) L_{\neq}(f) + (\tau) L_{=}(f)$. This states that the contrastive loss is the sum of the loss suffered when the negative and positive pair come from different classes ($L_{\neq}(f)$) as well as the loss when they come from the same class ($L_{=}(f)$). In an ideal setting ($L_{=}(f)$) would approach 0, but this is impossible without direct access to the underlying class distribution $\rho$. However, it may be the case that there exists another modality of data during training that provides us with a distribution $\rho_{clin}$ with the property that the $KL(\rho_{clin}||\rho) \leq \epsilon$, where $\epsilon$ is sufficiently small. In this case, the $D_{sim}$ and $D_{neg}$ could be drawn from $\rho_{clin}$ in the form:  $D_{sim} = \displaystyle \mathop{\mathbb{E}}_{c \sim \rho_{clin}} D_{c}(x)D_{c}(x^+)$ and $D_{neg} = \displaystyle \mathop{\mathbb{E}}_{c \sim \rho_{clin}} D_{c}(x^-)$. If $\rho_{clin}$ is a sufficiently good approximation for $\rho$, then this has a higher chance for the contrastive loss to choose positives and negatives from different class distributions and have an overall lower resultant loss. 

In this work, this related distribution that is in excess comes from the availability of clinical information within the unlabeled data and acts to form the $\rho_{clin}$ that we can use for choosing positives and negatives. As discussed in the introduction, this clinical data acts as a surrogate for the true distribution $\rho$ that is based on the severity of disease within the dataset and thus has the theoretical properties discussed. We also consider that there may exist many possible $\rho_{clin} \in P_{clin}$ where $P_{clin}$ is the set of all possible clinical distriubtions. In our case, these clinical distributions can come from the clinical values of BCVA, CST, and Eye ID which form the distributions $\rho_{bcva}$, $\rho_{cst}$, and $\rho_{eyeid}$. Additionally, we further show how these distributions can be utilized in tandem with each other to create distributions of the form $\rho_{bcva+cst}$, $\rho_{bcva+eye}$, $\rho_{cst+eye}$ and $\rho_{bcva+cst+eye}$. This builds and expands on previous work that only consider the distribution $\rho_{patientid}$. 

\section{Related Works}
\subsection{Clinical Data and Contrastive Learning}
The general idea of contrastive learning is to teach the model an embedding space where similar pairs of images project closer together and dissimilar pairs of images are projected apart. Approaches such as \cite{chen2020simple,he2020momentum,caron2020unsupervised, grill2020bootstrap} all generate similar pairs of images through various types of data augmentations such as random cropping, multi-cropping, and different types of blurs and color jitters. A classifier can then be trained on top of these learned representations while requiring fewer labels for satisfactory performance. The authors in~\cite{prabhushankar2021contrastive} augment contrastive class-based gradients and then train a classifier on top of the existing network. Other work \cite{alaudah2019structure, alaudah2018learning} used a contrastive learning setup with a similarity retrieval metric for weak segmentation of seismic structures. \cite{kokilepersaud2022volumetric} used volumetric positions as pseudo-labels for a supervised contrastive loss. Hence, contrastive learning presents a way to utilize a large amount of unlabeled data for performance improvements on a small amount of labeled data.
 
 The literature on self-supervised learning has shown that it is possible to leverage data augmentations as a means to create positive pairs for a contrastive loss. As discussed in the introduction, this isn't so simple within the medical domain due to issues with the diversity of data and small regions corresponding to important biomarkers. Previous work has shown that it is possible to use contrastive learning with augmentations on top of an Imagenet~\cite{deng2009imagenet} pretrained model to improve classification performance for x-ray biomarkers \cite{sowrirajan2020moco}. However, this is sub-optimal in the sense that the model required supervision from a dataset with millions of labeled examples. As a result, recent work has explored the idea of using medically consistent meta-data as a means of finding positive pairs of images alongside augmentations for a contrastive loss function. \cite{azizi2021big} showed that using images from the same medical pathology as well as augmentations for positive image pairs could improve representations beyond standard self-supervision. \cite{chen2021disease} demonstrated utilizing contrastive learning with a transformer can learn embeddings for electronic health records that can correlate with various disease concepts. Similarly, \cite{zhang2020contrastive} utilized pairings of images from x-rays with their textual reports as a means of learning an embedding for classification of various chest x-ray biomarkers. \cite{vu2021medaug,azizi2021big,diamant2021patient} investigated choosing positive pairs from images that exist from the same patient for the tasks of x-ray feature detection and ECG modeling. \cite{liang2021contrastive} used a contrastive loss to align textual and image embeddings within a chest x-ray setting. \cite{wang2020contrastive} incorporated a contrastive loss to align embeddings from different distributions of CT scans.  These works demonstrate the potential of utilizing clinical data within a contrastive learning framework. However, these methods were tried on limited clinical data settings, such as choosing images from the same patient or position relative to other tissues. Our work builds on these ideas by explicitly using measured clinical labels from an eye-disease setting as its own label for training a model. We also show the impact that drawing from multiple different clinical label distributions has on performance. Additionally, we perform our experiments in a novel setting with varying biomarker granularity to study the impact that this has on traditional algorithms. By doing this, we present a comprehensive assessment of what kinds of clinical data can possibly be used as well as how this clinical data can be used in tandem with each other to create a more robust representation space.

\subsection{Deep Learning and OCT}
 A desire to improve timely accurate diagnosis have led to applying deep learning ideas into detecting pathologies and biomarkers directly from OCT slices of the retina. Early work involved a binary classification task between healthy retina scans and scans containing age-related macular degeneration \cite{lee2017deep}. \cite{temel2019relative} introduced a technology to do relative afferent pupillary defect screening through a transfer learning methodology. \cite{kermany2018identifying} showed that transfer learning methods could be utilized to classify OCT scans based on the presence of key biomarkers. \cite{logan2022multimodal} showed how a dual-autoencoder framework with physician attributes could improve classification performance for OCT biomarkers. Subsequent work from \cite{schlegl2018fully} showed that semantic segmentation techniques could identify regions of fluid that are oftentimes indicators of different diseases. \cite{de2018clinically} expanded previous work towards segmentation of a multitude of different biomarkers and connected this with referral for different treatment decisions. \cite{pekala2019deep} showed that segmentation could be done in a fine-grained process by separating individual layers of the retina. Other work has demonstrated the ability to detect clinical information from OCT scans which is significant for suggesting correlations between different domains. \cite{kawczynski2020development} showed that a model trained entirely on OCT scans could learn to predict the associated BCVA value. Similarly \cite{arcadu2019deep} showed that values such as retinal thickness could be learned from retinal fundus photos. 
 
  Within the context of self-supervised algorithms,  \cite{rivail2019modeling} introduced a novel pretext task that involved predicting the time interval between OCT scans taken by the same patient. \cite{zhang2021twin} showed how a combination of different pretext tasks such as rotation prediction and jigsaw re-ordering can improve performance on an OCT anomaly detection task. \cite{qiu2019self} showed how assigning pseudo-labels from the output of a classifier can be used to effectively identify labels that might be erroneous. These works all identify ways to use variants of deep learning to detect important biomarkers in OCT scans. However, they differ fundamentally from our work in the sense that they don't present a framework to integrate clinical data within a contrastive learning algorithm.

 \subsection{Contrastive Loss Functions}

 Traditional contrastive learning algorithms \cite{chen2020simple,li2020prototypical,chen2020improved} make use of the information theoretic noise contrastive estimation loss (Info-NCE) that takes the form:

  $$
    L_{self} = -\sum_{i\in{I}} log\frac{exp(z_{i}\cdot z_{j(i)}/\tau)}{\sum_{a\in{A(i)}}exp(z_{i}\cdot z_{a}/\tau)}
$$
This loss takes as input an image $i$ into an encoder network $f(\cdot)$ to produce an output $r_{i}$. This is passed through a projection network to produce an embedding $z_{i}$. The loss is trained to maximize the dot product between its embedding $z_{i}$ and the embedding of its data augmentation $z_{j(i)}$ while maximizing the distance between all other embeddings in the batch that are represented by the set $a \in A(i)$ with the notation $z_{a}$. Note that because the loss uses an image and its augmentation as its positives, its completely self-supervised in the manner in which it does its feature extraction. It does not have the capability to incorporate annotation information in order to choose a better set of positives and negatives. This changed with the introduction of the supervised contrastive loss \cite{khosla2020supervised}. The introduced loss function can be represented by: 
 $$
    L_{supcon} = \sum_{i\in{I}} \frac{-1}{|P(i)|}\sum_{p\in{P(i)}}log\frac{exp(z_{i}\cdot z_{p}/\tau)}{\sum_{a\in{A(i)}}exp(z_{i}\cdot z_{a}/\tau)}
$$
 The objective behind this function is to enforce similarity between images with the same label and dissimilarity between images that have differing labels. Using the language of contrastive learning, this means that labels are used to identify the positive and negative pairs, rather than augmentations. The loss is computed on each image $x_{i}$ where $i \in{I}  = {1, ..., 2N}$ represents the index for each instance within the overall augmented batch. Each image $x_{i}$ is passed through an encoder network $f(\cdot)$, producing a lower dimensional representation $r_{i}$. This vector is further compressed through a projection head to produce the embedding vector $z_{i}$. Positive instances for image $x_{i}$ come from the set $P(i)$ and all positive and negative instances come from the set $A(i)$. The loss function operates in the embedding space where the goal is to maximize the cosine similarity between embedding $z_{i}$ and its set of positives $z_{p}$. $\tau$ is a temperature scaling parameter.

 Our work differs from this loss in the sense that we define images belonging to the same class through the use of clinical labels. In this sense, our proposed loss framework can be described as a clinically aware supervised contrastive loss. Additionally, previous work has not studied how the supervised contrastive loss can be used as part of a linear combination of different label distributions. We propose such a method and show how this combined loss leads to more robust performance metrics.

 \subsection{OCT Datasets}
 Previous OCT datasets for machine learning have labels for specific segmentation and classification tasks regarding various retinal biomarkers and conditions. \cite{kermany2018labeled} contains OCT scans for 4 classes of OCT disease states: Healthy, Drusen, DME, and choroidal neovascularization (CNV). \cite{farsiu2014quantitative} and \cite{melinvsvcak2021annotated} introduced OCT datasets for segmentation of regions with age-related macular degeneration (AMD). \cite{chiu2015kernel} created a dataset for segmentation of regions with DME. In all cases, these datasets do not come with associated comprehensive clinical information nor a wide range of biomarkers to be detected. Recently, the \texttt{OLIVES} \cite{prabhushankar2022olives} dataset was introduced that enabled many of the experiments in this paper, through their introduction of associated clinical labels and biomarkers of varying difficulty and granularity. 

 \begin{figure}[ht]
\centering
\includegraphics[width = \columnwidth]{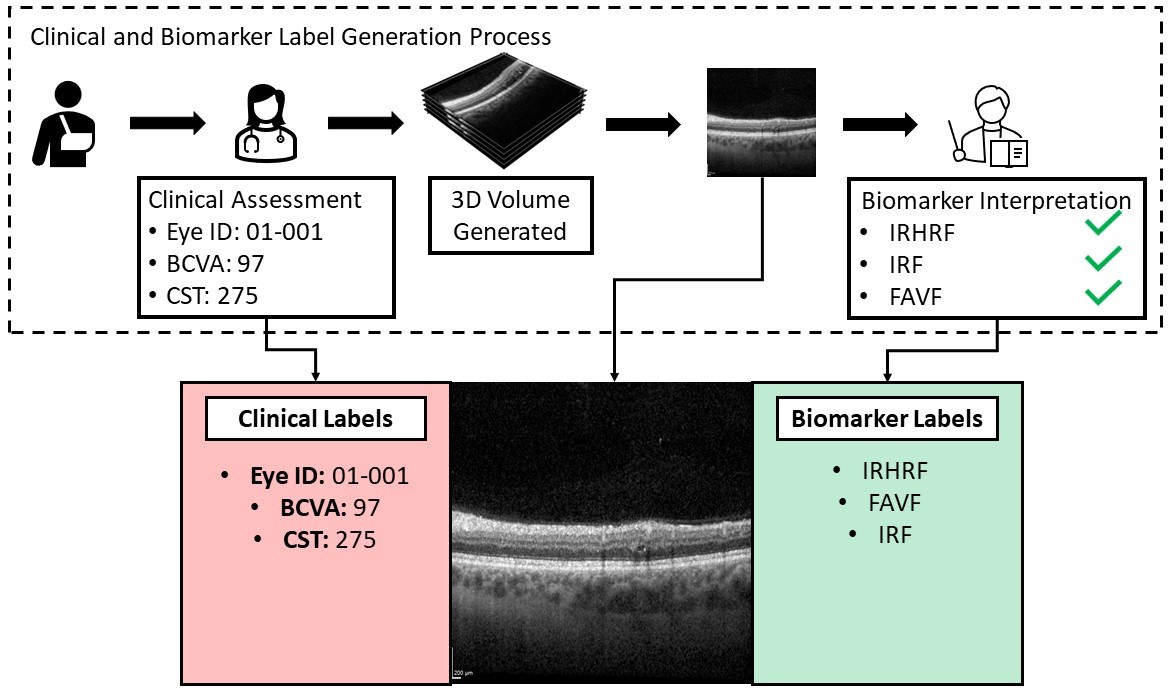}
\caption{ This is an illustration of the association between a single OCT scan and both clinical labels and biomarker labels. These labels are obtained at various stages during the healthcare process. }
\label{fig: image_single_example}
\end{figure}

\section{Methodology}

\subsection{Dataset}

\begin{figure*}[ht]
\centering
\includegraphics[scale = .4]{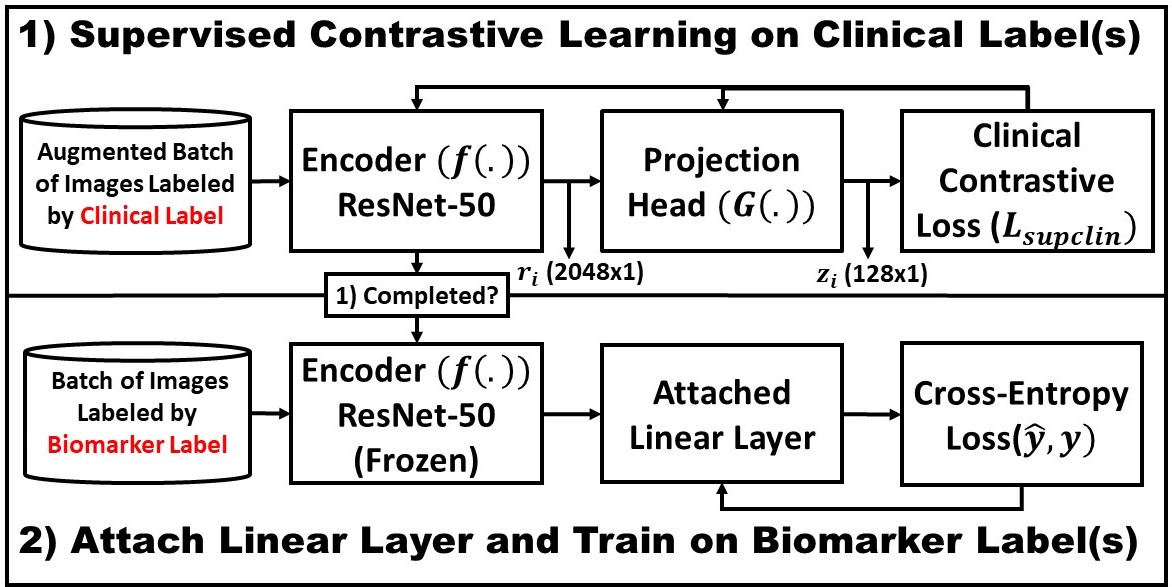}
\caption{ Overview of Experimental Setup. 1) Supervised contrastive learning is performed on the larger amount of available clinical data provided in OLIVES Clinical. 2) This trained encoder then has its weights frozen and a linear classifier is trained utilizing the smaller amount of data from the OLIVES Biomarker dataset.}
\label{fig: supcon}
\end{figure*}
 We make use of the \texttt{OLIVES} \cite{prabhushankar2022olives} dataset for our studies. This dataset is composed of 78108 OCT scans from two clinical trials. Every image is associated withe the clinical information of Eye identity, BCVA, and CST that was naturally collected during the process of patient treatment during the clinical trial. A smaller subset of 9408 images was additionally labeled by a trained grader for the presence or absence of 16 different biomarkers present in each image.   In addition to this information provided by the studies, a trained grader performed interpretation on OCT scans for the presence of 20 different biomarkers including: Intra-Retinal Hyper-Reflective Foci (IRHRF), Partially Attached Vitreous Face (PAVF), Fully Attached Vitreous Face (FAVF), Intra-Retinal Fluid (IRF), and Diffuse Retinal Thickening or Macular Edema (DRT/ME). The trained grader was blinded to clinical information whilst grading each of 49 horizontal SD-OCT B-scans of both the first and last study visit for each individual eye. This results in a subset of both studies now having just clinical labels and a smaller subset having both clincial and biomarker labels as observed in Figure \ref{fig:quantity}. This can be further understood by Figure \ref{fig: image_single_example} where we can see that, for a single image, a wide variety of both clinical and biomarker labels can exist.  Open adjudication was done with an experienced retina specialist for difficult cases.

As a result, for each OCT scan labeled for biomarkers, there exists a one-hot vector indicating the presence or absence of 16 different biomarkers. Not all of these 16 biomarkers exist in sufficiently balanced quantities to train a model to identify their presence or absence within an image. Hence, we use Intraretinal Hyperreflective Foci (IRHRF), Partially Attached Vitreous Face (PAVF), Fully Attached Vitreous Face (FAVF), Intraretinal Fluid (IRF), and Diffuse Retinal Thickening or Diabetic Macular Edema (DRT/ME) as the biomarkers that are studied in this paper. These biomarkers were also chosen in order to enable a novel granularity analysis as discussed in the introduction. From Figure \ref{fig: examples} we observe that the biomarkers of IRF and DME exhibit low granularity in the sense that they are easily distinguishable, while the biomarkers IRHRF, PAVF, and FAVF are features with high granularity that are hard to rectify from the rest of the image.  It should be noted that because OLIVES dataset is composed of images from two separate trials with different intended outcomes, this results in each having different types of clinical information. When combining the trials together, we only focus on the clinical data that is commonly held by both trials: BCVA, CST, and Eye ID.  We present the manifestation of distributions within the OLIVES dataset for the clinical values of BCVA, CST, and Eye ID in Figures \ref{fig:patient_dist}, \ref{fig:bcva_dist}, and \ref{fig:cst_dist}. It can be observed that the distribution of each value image quantities isn't noticeably biased towards any specific value. Instead, for each value, there is diversity in terms of the number of different eyes and number of images.

In total, the \texttt{OLIVES} dataset provides data from 96 unique eyes and 87 unique patients. We take 10 unique eyes from both clinical trials in the \texttt{OLIVES} dataset and use the data from these 20 eyes to create the test set. The remaining 76 eyes are utilized for training in all experiments. Since the objective is to evaluate the model's performance in identifying each biomarker individually, a test set for each biomarker is created. This is done by randomly sampling 500 images with the biomarker present and 500 images with the biomarker absent from the data associated with the test eyes. In this way, we ensure that the test set for each individual biomarker is balanced.

In the case of nine patients there exists data for the left and right eye. This leads to questions as to whether this may potentially bias any experiments that are performed by having both in the dataset. However, in the original clinical trials on which \texttt{OLIVES} is based, each eye is treated as independent from every other eye and different experimental settings are applied to each. Consequently, random splits of experiments were made on the basis of the eye identity, rather than the patient identity. This also means that data is recorded for individual eyes, rather than individual patients. So as to align with the setup of the original studies, we make our train and test split on the basis of the eye identity. Ablation studies later in the paper show that our methods perform comparably even if we were to make the split based solely on patient identity.

\begin{table*}[]
\centering
\begin{tabular}{@{}cccccc@{}}
\toprule
\multicolumn{6}{c}{Individual Biomarker Performance (Accuracy / F1-Score)}   
\\ \midrule
\multicolumn{1}{|c|}{Method} & \multicolumn{1}{c|}{IRF} & \multicolumn{1}{c|}{DME} & \multicolumn{1}{c|}{IRHRF} & \multicolumn{1}{c|}{FAVF} & \multicolumn{1}{c|}{PAVF} \\\midrule
PCL  \cite{li2020prototypical} &  \hlgreen{$74.45\% \pm.212 / .687$} & \hlgreen{$78.74\% \pm.510 / .741$} & \hlgreen{$56.36\% \pm.404 / .663$} &  \hlgreen{$53.43\% \pm.057 / .635$} & \hlgreen{$50.06\% \pm.305 / .028$}              \\

SimCLR \cite{chen2020simple}  &  \hlgreen{$74.16\% \pm.115 / .689$} & \hlgreen{$79.44\% \pm.154 / .758$} & \hlgreen{$61.53\% \pm1.19 / .679$} &  \hlgreen{$73.53\% \pm.305 / .739$} & \hlgreen{$53.90\% \pm.519 / .269$}                  \\
Moco v2 \cite{chen2020improved} &  \hlgreen{$\textbf{76.23}\% \pm.208 / .721$} & \hlgreen{$77.89\% \pm.101 / .724$} & \hlgreen{$57.20\% \pm.360 / .678$} &  \hlgreen{$64.70\% \pm.458 / .696$} & \hlgreen{$51.80\% \pm.400 / .119$}                 \\ \bottomrule

Eye ID  &  \hlgreen{$73.20\% \pm.100 / .692$} & \hlgreen{$77.58\% \pm.268 / .740$} & \hlgreen{$58.90\% \pm.529 / .681$} &  \hlgreen{$82.00\% \pm.100 / .841$} & \hlgreen{$\textbf{68.53}\% \pm.208 / \textbf{.604}$ }                \\
\hline
CST   &  \hlgreen{$73.83\% \pm.057 / .670$} & \hlgreen{$78.66\% \pm.058 / .736$} & \hlgreen{$61.33\% \pm.057 / .676$} &  \hlgreen{$79.23\% \pm.351 / .797$} & \hlgreen{$57.96\% \pm.152 / .372$}                      \\

BCVA  &  \hlgreen{$74.13\% \pm.152 / .689$} & \hlgreen{$80.32\% \pm.101 / .766$} & \hlgreen{$\textbf{63.83}\% \pm.321 / \textbf{.692}$} &  \hlgreen{$78.90\% \pm.529 / .800$} & \hlgreen{$60.93\% \pm.208 / .452$}   \\

BCVA + Eye ID &  \hlgreen{$73.30\% \pm.435 / .701$} & \hlgreen{$80.02\% \pm.101 / .775$} & \hlgreen{$55.93\% \pm.473 / .679$} &  \hlgreen{$\textbf{82.56}\% \pm.305 / \textbf{.843}$} & \hlgreen{$66.16\% \pm.115 / .561$}             \\

BCVA + CST  &  \hlgreen{$74.26\% \pm.208 / .695$} & \hlgreen{$80.22\% \pm.101 / .765$} & \hlgreen{$62.00\% \pm.400 / .681$} &  \hlgreen{$81.00\% \pm.100 / .818$} & \hlgreen{$60.66\% \pm.416 / .447$}   \\              
CST + Eye ID  &  \hlgreen{$75.66\% \pm.152 / \textbf{.728}$} & \hlgreen{$\textbf{80.86}\% \pm.154 / \textbf{.786}$} & \hlgreen{$59.13\% \pm.208 / .686$} &  \hlgreen{$80.60\% \pm.200 / .825$} & \hlgreen{$60.53\% \pm.551 / .436$}            \\

BCVA + CST + Eye ID  &  \hlgreen{$74.26\% \pm.416 / .697$} & \hlgreen{$80.15\% \pm.058 / .765$} & \hlgreen{$61.43\% \pm.305 / .692$} &  \hlred{$82.33\% \pm.057 / .837$} & \hlgreen{$62.16\% \pm.057 / .492$}        \\

 \bottomrule
\end{tabular}
\caption{This table shows the performance of each contrastive training strategy in terms of accuracy and F1-Score for each individual biomarker used in this study. We also perform a significance test for the best result associated with each biomarker. This was done by running 3 different trials for every method on different seeds to produce a population of model outputs associated with each method. We then perform a pairwise t-test between the population of outputs for the best performing model and every other model in the column to compare whether the means of the output performance for the two groups are significantly different from each other. If the p-value generated by this test was less than .05 we deemed the performance difference between the best model and the tested model as significant and highlighted it as green in the table above.}
\label{tab:individual_biomarker_table}
\end{table*}
\subsection{Overall Framework}

The overall block diagram of the proposed method is summarized in Figure \ref{fig: supcon}. Within the OLIVES Clinical dataset, each individual image is associated with the clinical values BCVA, CST, and Eye ID that are taken during the original patient visit. For each experiment, we first choose one of these clinical values to act as a label for each image in the dataset. 

Given an input batch of data, $(x_{k}$, and clinical label, $(y_{k}$, pairs $(x_{k},y_{k})_{k=1,...,N}$, we perform augmentations on the batch twice in order to get two copies of the original batch with $2N$ images and clinical labels. These augmentations are random resize crop to a size of 224, random horizontal flips, random color jitter, and data normalization. These are sensible from a medical perspective because they don't disrupt the general structure of the image.
This process produces a larger set $(x_{l},y_{l})_{l=1,...,2N}$ that consists of two versions of each image that differ only due to the random nature of the applied augmentation. Thus, for every image $x_{k}$ and clinical label $y_{k}$ there exists two views of the image $x_{2k}$ and $x_{2k-1}$ and two copies of the clinical labels that are equivalent to each other: $y_{2k-1} = y_{2k} = y_{k}$.

From this point, we perform the first step in Figure \ref{fig: supcon}, where supervised contrastive learning is performed on the identified clinical label. The clinically labeled augmented batch is forward-propagated through an encoder network $f(\cdot)$ that we set to be the ResNet-50 architecture \cite{he2016deep}. This results in a 2048-dimensional vector $r_{i}$ that is sent through a projection network $G(\cdot)$, which further compresses the representation to a 128-dimensional embedding vector $z_{i}$. $G(\cdot)$ is chosen to be a multi-layer perceptron network with a single hidden layer. This projection network is utilized only to reduce the dimensionality of the embedding before computing the loss and is discarded after training. A supervised contrastive loss is performed on the output of the projection network in order to train the encoder network. In this case, embeddings with the same clinical label are enforced to be projected closer to each other while embeddings with differing BCVA labels are projected away from each other.  Our introduced clinical supervised contrastive loss process can be understood by:
$$
    L_{clinical} = \sum_{i\in{I}} \frac{-1}{|C(i)|}\sum_{c\in{C(i)}}log\frac{exp(z_{i}\cdot z_{c}/\tau)}{\sum_{a\in{A(i)}}exp(z_{i}\cdot z_{a}/\tau)}
$$
where $i$ is the index for the image of interest $x_{i}$.
All positives $c$ for image $x_{i}$ are obtained from the set $C(i)$ and all positive and negative instances $a$ are obtained from the set $A(i)$. Every element $c$ of $C(i)$ represents all other images in the batch with the same clinical label $c$ as the image of interest $x_{i}$. Additionally, $z_{i}$ is the embedding for the image of interest, $z_{c}$ represents the embedding for the clinical positives, and $z_{a}$ represents the embeddings for all positive and negative instances in the set $A(i)$. $\tau$ is a temperature scaling parameter that is set to .07 for all experiments.  The loss function operates in the embedding space where the goal is to maximize the cosine similarity between embedding $z_{i}$ and its set of clinical positives $z_{c}$. It should be explicitly stated that the set $C(i)$ can represent any clinical label of interest. Throughout the rest of the paper we will use certain conventions to make the choice of clinical label in the loss transparent. For example, a loss represented as $L_{BCVA}$ indicates a supervised contrastive loss where BCVA is utilized as the clinical label of interest. Furthermore, it is also possible to create an overall loss that is a linear combination of several losses on different clinical labels. This can be represented by: 

$$L_{BCVA + CST} = L_{BCVA} + L_{CST}$$ 

where each clinical value respectively (BCVA and CST) acts as a label for its respective loss. In this way, we are creating a linear combination of losses from different clinical label distributions for the same image. 

After training the encoder with  clinical supervised contrastive loss, we move to the second step in Figure \ref{fig: supcon} where the weights of the encoder are frozen and a linear layer is appended to the output of the encoder. This setup is trained on the biomarker(s) of interest. This linear layer is trained using cross-entropy loss to distinguish between the presence or absence of the biomarker(s) of interest in the OCT scan. In this way, we leverage knowledge learnt from training on clinical labels to improve performance on classifying biomarkers. The previously trained encoder with the supervised contrastive loss on the clinical label from step 1 produces the representation for the input and this representation is fine-tuned with the linear layer to distinguish whether or not the biomarker(s) of interest is present.

\begin{table}[]
\centering
\scalebox{.9}{
\begin{tabular}{@{}ccccc@{}}
\toprule
\multicolumn{5}{c}{Performance Metrics Averaged Across All Biomarkers}   
\\ \midrule
\multicolumn{1}{|c|}{Method} & \multicolumn{1}{c|}{AUROC} & \multicolumn{1}{c|}{Precision} & \multicolumn{1}{c|}{Sensitivity} & \multicolumn{1}{c|}{Specificity}\\\midrule
PCL  \cite{li2020prototypical}  & $.676 \pm .002$ &  .676 & .572 & .681            \\

SimCLR \cite{chen2020simple}  & $.761 \pm .003$ &  .748 & .591 & .779                  \\
Moco v2 \cite{chen2020improved} & $.737 \pm .002$ &  .734 & .597 & .711                 \\ \bottomrule

Eye ID   & $.802 \pm .001$ &  .769 & .701 & .741                  \\ \hline
CST      & $.793 \pm .001$ &  \textbf{.792} & .601 & \textbf{.803}                   \\
BCVA     & $.801 \pm .001$ &  .785 & .640 & .792                 \\

BCVA + Eye ID   & $.804 \pm .002$ &  .756 & \textbf{.723} & .708           \\

BCVA + CST  & $.807 \pm .001$ &  .783 & .643 & .789               \\
CST + Eye ID  & $\textbf{.819} \pm .001$ &  .756 & .694 & .732            \\

BCVA + CST + Eye ID & $.817 \pm .001$ &  .776 & .677 & .764        \\

 \bottomrule
\end{tabular}
}
\caption{In this table, we obtain the AUROC, Precision, Sensitivity, and Sensitivity across all 5 biomarkers for each contrastive method and then average them together to get a sense of performance across all biomarkers.}
\label{tab:all_biomarker_table}
\end{table}

\section{Experiments}

\subsection{Training Details}
Care was taken to ensure that all aspects of the experiments remained the same whether training was done via supervised or self-supervised contrastive learning on the encoder or cross-entropy training on the attached linear classifier. The encoder utilized was kept as a ResNet-50 architecture. The applied augmentations are random resize crop to a size of 224, random horizontal flips, random color jitter, and data normalization to the mean and standard deviation of the respective dataset. The batch size was set at 128. Training was performed for 25 epochs in every setting. A stochastic gradient descent optimizer was used for contrastive pre-training with a learning rate of .05, weight decay of .0001, and momentum of .9. The same chosen hyper-parameters were used during cross-entropy fine-tuning except the learning rate was changed to .001 for stability during this step. The comparison methods of SimCLR\cite{chen2020simple}, Moco v2\cite{chen2020improved}, and PCL \cite{li2020prototypical} were trained in the same manner with certain hyper-parameters specific to each method. Specifically, Moco v2 was set to its default queue size of 65536. Additionally, PCL has hyper-parameters specific to its clustering step, but the original documentation made these parameters specific to the Imagenet \cite{deng2009imagenet} dataset on which it was originally built for. To fit these parameters to our setting, the clustering step was reduced in size.

\subsubsection{Metrics and Notation}
During supervised contrastive training, a choice of a single clinical parameter or combination of parameters is chosen to act as labels. For example, in Table \ref{tab:individual_biomarker_table} when the method is specified as BCVA this indicates a supervised contrastive loss $L_{BCVA}$ where BCVA is utilized as the label of interest for the images in the dataset. Additionally, BCVA + CST refers to a linear combination of supervised contrastive losses that can be expressed as $L_{BCVA+CST} = L_{BCVA} + L_{CST}$ where each clinical value respectively  acts as a label for its respective loss. A linear layer is then appended to this trained encoder and trained on the biomarker labels present, which consists of approximately 7,500 images during training. This linear layer is trained on each biomarker individually and accuracy as well as F1-score in detecting the presence of each individual biomarker is reported. Any table that reports the standard average area under the receiver operating curve (AUROC) is reporting the AUROC calculated after finding the AUROC for each individual biomarker test set and then averaging them together in order to get a single metric summarizing performance. The same is done to get an average precision, sensitivity, and specificity. Performance is also evaluated in a multi-label classification setting where the goal is to correctly identify the presence or absence of all 5 biomarkers at the same time. This is evaluated using averaged AUROC over all 5 classes within this multi-label setting. Every reported value is the average of 3 runs of the specified method and the standard deviation for both accuracy and AUROC are reported.
\begin{figure}[t]
\centering
\includegraphics[scale = .4]{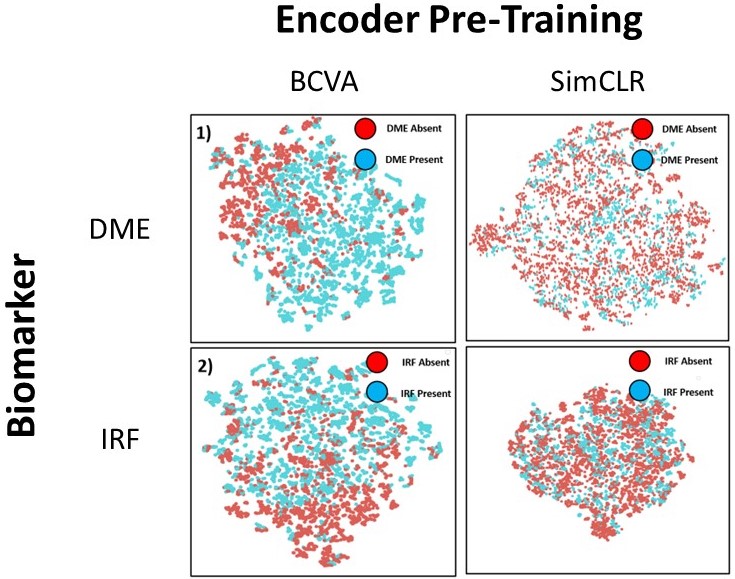}
\caption{Images from the OLIVES Biomarker dataset labeled by the presence or absence of DME and Fluid IRF were fed into an encoder network trained by using BCVA values as the label as well as a SimCLR strategy. This produced an embedding for each image. These embeddings were visualized using t-SNE \cite{van2008visualizing}  with two components. It can be observed that from an encoder trained using BCVA labels with the supervised contrastive loss, we can effectively achieve an embedding space that is separable with respect to biomarkers while the standard contrastive learning method shows no separability for either of the biomarkers. }
\label{fig: dme_ebed}
\end{figure}

\begin{table}[]
\centering
\scalebox{.8}{
\begin{tabular}{@{}ccccc@{}}
\toprule
\multicolumn{5}{c}{Averaged AUROC Across Different Splits Of Patients}   
\\ \midrule
\multicolumn{1}{|c|}{Method} & \multicolumn{1}{c|}{PS1} & \multicolumn{1}{c|}{PS2} & \multicolumn{1}{c|}{PS3} & \multicolumn{1}{c|}{Average}\\\midrule
PCL  \cite{li2020prototypical}  &   $.676 \pm .002$ & $.759 \pm .002$ & $.779 \pm .002$ & \hlgreen{$.738 \pm .054$}            \\

SimCLR \cite{chen2020simple}     &   $.761 \pm .003$ & $.770 \pm .001$ & $.795 \pm .001$ & \hlgreen{$.775 \pm .018$}               \\
Moco v2 \cite{chen2020improved}  &   $.737 \pm .002$ & $.769 \pm .002$ & $.729 \pm .002$ & \hlgreen{$.745 \pm .021$}                \\ \bottomrule

Eye ID   &   $.802 \pm .001$ & $.782 \pm .001$ & $ .801 \pm .001$ & \hlgreen{$.795 \pm .006$}                  \\ \hline
CST     &   $.793 \pm .001$ & $.794 \pm .001$ & $.783 \pm .001$ & \hlgreen{$.790 \pm .006$}                    \\
BCVA    &   $.801 \pm .001$ & $.788 \pm .001$ & $.776 \pm .001$ & \hlgreen{$.788 \pm .012$}                \\

BCVA + Eye ID  &   $.804 \pm .002$ & $.796 \pm .001$ & $.821 \pm .001$ & \hlgreen{$.807 \pm .013$}            \\

BCVA + CST  &   $.807 \pm .001$ & $.795 \pm .001$ & $.797 \pm .001$ & \hlgreen{$.800 \pm .007$}               \\
CST + Eye ID  &   $\textbf{.819} \pm .001$ & $\textbf{.833} \pm .001$ & $\textbf{.828} \pm .001$ & \hlgreen{$\textbf{.827} \pm .007$}             \\

BCVA + CST + Eye ID  &   $.817 \pm .001$ & $.792 \pm .001$ & $.821 \pm .001$ & \hlgreen{$.810 \pm .015$}        \\

 \bottomrule
\end{tabular}}
\caption{We show performance in terms of average AUROC across all biomarkers for different training and testing splits for patient groups. PS1, PS2, and PS3 refer to the three different patient splits created. The average column refers to the average performance across all the splits. We also performed a statistical significance test over the resultant average performance. A green highlight indicates that the result is significant with respect to the best result on all patient splits and red indicates that the result lacked statistical significance. The threshold used to determine significance is a p-value of .05. }
\label{tab:patient_split_table}
\end{table}

\begin{table}[]
\centering
\begin{tabular}{@{}cc@{}}
\toprule
\multicolumn{2}{c}{Averaged AUROC on Smaller Contrastive Subset}   
\\ \midrule
\multicolumn{1}{|c|}{Method} & \multicolumn{1}{c|}{AUROC} \\\midrule
PCL  \cite{li2020prototypical}     & $.667 \pm .006$         \\

SimCLR \cite{chen2020simple}       & $.689 \pm .002$             \\
Moco v2 \cite{chen2020improved}    & $.559 \pm .002$              \\ \bottomrule

Eye ID        & $.748 \pm .001$             \\ \hline
CST          & $.781 \pm .001$               \\
BCVA               & $.775 \pm .002$          \\

Leakage Index        & $\textbf{.801} \pm .001$      \\

DRSS        & $.618 \pm .002$         \\
Diabetes Type    & $.646 \pm .007$          \\

Years with Diabetes  & $.761 \pm .001$      \\
Gender  & $.609 \pm .003$      \\

 \bottomrule
\end{tabular}
\caption{To study the impact of reducing the amount of available training data during the contrastive training set, we utilized the PRIME subset of the OLIVES dataset for training the ResNet-50 network. In this case, we have access to additional clinical values that we can observe because this subset has additional clinical information that can be investigated. We use the average AUROC across all biomarkers as the metric of interest.}
\label{tab:clinical_diverse_table}
\end{table}

\begin{table}[]
\centering
\begin{tabular}{@{}ccc@{}}
\toprule
\multicolumn{3}{c}{Averaged AUROC on Different Sized Architectures}   
\\ \midrule
\multicolumn{1}{|c|}{Method} & \multicolumn{1}{c|}{R-18} & \multicolumn{1}{c|}{R-50} \\\midrule
PCL  \cite{li2020prototypical}     & $.716 \pm .002$ & $.676 \pm .002$         \\

SimCLR \cite{chen2020simple}      & $.719 \pm .002$ & $.761 \pm .003$              \\
Moco v2 \cite{chen2020improved}    & $.748 \pm .002$ & $.737 \pm .002$              \\ \bottomrule

Eye ID        & $.771 \pm .001$ & $.802 \pm .001$             \\ \hline
CST          & $.771 \pm .003$ & $.793 \pm .001$               \\
BCVA          & $.753 \pm .002$ & $.801 \pm .001$               \\

BCVA + Eye ID    & $.792 \pm .003$ & $.804 \pm .002$          \\

BCVA + CST      & $.796 \pm .002$ & $.807 \pm .001$           \\
CST + Eye ID   & $.794 \pm .004$ & $\textbf{.819} \pm .001$          \\

BCVA + CST + Eye ID  & $ \textbf{.816} \pm .004$ & $.817 \pm .001$      \\

 \bottomrule
\end{tabular}
\caption{This table shows the average AUROC metric across all biomarkers for different sized architectures. R-18 and R-50 refers to ResNet-18 and ResNet-50 \cite{he2016deep} respectively.}
\label{tab:arch}
\end{table}
\subsection{Biomarker Detection Experiments}
We first evaluate the capability of our method to leverage a larger amount of clinical labels for performance improvements on the smaller biomarker subset. This involves supervised contrastive training of the encoder network on the training set with clinical labels, consisting of approximately 60,000 images. Tables \ref{tab:individual_biomarker_table} and \ref{tab:all_biomarker_table} shows that applying our method of choosing positives and negatives based on some clinical label or combination of labels leads to improvements on classification accuracy and f1-score of each biomarker individually as well as an improved average AUROC, precision, sensitivity, and specificity over all 5 biomarkers when compared against state of the art self-supervised algorithms. We can also observe visually how well training on a clinical label performs in creating a separable embedding space for biomarkers through Figure \ref{fig: dme_ebed}. This figure was generated by taking a model trained with a supervised contrastive loss on BCVA values as well as a model trained with the SimCLR framework and inputting the test set labeled by the biomarkers DME and Intraretinal Fluid. This produces embeddings for each image in the test set. These embeddings are projected into a lower dimensional space with 2-D t-SNE \cite{van2008visualizing} algorithm. It can be observed that the resulting representation can separate between present and absent forms of DME and Fluid IRF without having explicit training for these labels. However, the model trained with just self-supervision shows almost no separability with respect to these biomarkers. This acts as validation to the relationship between both the clinical labels and biomarkers and gives insight into the improved results that we observe.  

Other interesting conclusions can be derived from observing the performance of the standard self-supervised methods compared to our clinically driven concepts. It is  observed in Table \ref{tab:individual_biomarker_table} that the self-supervised algorithms are comparable to our methods for IRF and DME, but are worse for IRHRF, FAVF, and PAVF. In other words, the traditional contrastive learning algorithms did comparably well on biomarkers that were easy to distinguish from the rest of the image, but did poorly on biomarkers that exhibited a higher degree of granularity.
This difference can be understood through the analysis performed by the authors of \cite{cole2021does}. In this work, the authors showed that the performance gap between contrastive learning and supervised learning increases as the granularity of the task correspondingly increases. They hypothesized that the contrastive loss tends to cluster images based on coarse-visual similarity. This is because the contrastive loss relies on creating positive instances of similar images from augmentations taken from an individual image. In this sense, there is a dependence on the individual image itself to have enough distinguishable features such that a contrast can be created with the negative instances in the loss function. This may not always be the case, especially when it comes to the medical domain where images can be very similar with the exception of small localized regions. We hypothesize that our method is better able to overcome this issue by providing an effective method to identify positive instances that are correlated through having similar clinical metrics. Instead of an over-reliance on augmentations from a single image, a more robust set of positive pairs can be found that allows the model to more effectively identify fine-grained features of OCT scans.  This is due to having a wider and more informative set of features to treat as positive instances, thus allowing the model to better identify distinguishing features when using a contrastive loss even for more fine-grained cases. 

Another aspect of the results in Table \ref{tab:individual_biomarker_table} is how well the used clinical labels correspond with the biomarker classification performance. In all cases, the results act as validation to the hypothesis that taking advantage of correlations that exist with certain clinical labels is beneficial for biomarker detection of individual OCT scans. However, from a medical perspective, certain outcomes would intuitively be more likely. For example, for IRF and DME it makes sense that the best performance is associated with using CST values because CST tends to more closely increase or decrease, depending on the severity of IRF and DME. In general, further medical insight is needed to determine what is expected in terms of the degree of correlations between biomarkers and individual clincal values.

\subsection{Robustness Experiments}

In this section, we investigate the performance of our algorithms in a variety of settings meant to test how robust our method is to various perturbations in the original setup. This includes different training and test splits based on different subsets of the patient pool, experiments where we reduce the amount of clinical and biomarker training data, studies on different sized backbone architectures, and experiments within the tougher setting of multi-label classification. In all experiments we summarize performance as the average of the AUROC found on each biomarker test set. Through these experiments we arrive at the somewhat surprising finding that using a combination of clinical labels in a linear combination of supervised contrastive losses actually out-performs all other methods whether self-supervised or on individual clinical labels. 

In our first experiment, we create two new training and test sets based on splitting by patient identities. We show in Table \ref{tab:patient_split_table} that our methods work even in the case where splits between training and testing are made on the basis of different patient identities than the one we originally started with. When we take the average across all patient identity splits, it should be noted that methods that utilized a combined loss such as CST + Eye ID were able to maintain a consistently higher performance across different splits, while maintaining a much lower variance across splits. This difference is further highlighted when comparing against the self-supervised algorithms which exhibited not only lower performance, but also a greater variance across the different splits. All of this indicates that these combined losses have a greater robustness with respect to whichever training and testing pool is utilized. A possible reason for this is that each clinical value can be thought of being associated with its own distribution of images. By having a linear combination of losses on two clinical values, we are effectively choosing positive instances from closely related, but slightly varying distributions. These distributions are visualized in Figures \ref{fig:bcva_dist} and \ref{fig:cst_dist}. It can be observed that because BCVA and CST have different ranges of potential values, as observed along the x-axis of this figure, this means that for any individual value there is a different number of associated eyes and images. Effectively, this means that there is varying diversity with respect to any individual label. This may allow the model to better learn features by allowing the model to sample from different proposal distributions that may better approximate the true disease severity distribution. 

We also observe the robustness of the combined clinical losses in the setting of different architecture choices. We see in Table \ref{tab:arch} that the combined clinical losses were the only ones that maintained consistently high performance across both the ResNet-18 and ResNet-50 architectures. Additionally, we see that in the case of self-supervised methods, PCL and SimCLR actually did worse on the larger architecture. This may be due to over-fitting on the easily distinguishable features while losing the ability to generalize to the higher granularity biomarkers. 

From a medical perspective, the success of our approach across patient, architecture, and eye splits is indicative of its generalizability. Fully self-supervised approaches are dependent on the data to have enough differentiating features such that augmentations can be used to learn useful features. In this sense, certain groups of patients will potentially be more informative than others, which limits the ability of these approaches to generalize. However, our approach, by choosing positives from a much wider set of data, does not have as strong of a dependence on the features present in any one image. From this perspecitve, there is a greater robustness in regards to patient distributions which corresponds to the better performance across all splits of data.

\begin{table}[]
\centering
\scalebox{.8}{
\begin{tabular}{@{}ccccc@{}}
\toprule
\multicolumn{5}{c}{Averaged Multi-Label AUROC with varying Biomarker Access}   
\\ \midrule
\multicolumn{1}{|c|}{Method} & \multicolumn{1}{c|}{25\%} & \multicolumn{1}{c|}{50\%} & \multicolumn{1}{c|}{75\%} & \multicolumn{1}{c|}{100\%}\\\midrule
Supervised & $.703 \pm .002$ & $.716 \pm .003$ & $.719 \pm .002$ & $.722 \pm .005$ \\

PCL  \cite{li2020prototypical}  &   $.675 \pm .003$ & $.681 \pm .004$ & $.683 \pm .002$ & $.681 \pm .002$            \\

SimCLR \cite{chen2020simple}     &   $.679 \pm .004$ & $.709 \pm .006$ & $.718 \pm .003$ & $.727 \pm .002$               \\
Moco v2 \cite{chen2020improved}  &   $.709 \pm .006$ & $.722 \pm .002$ & $.732 \pm .001$ & $.734 \pm .002$                \\ \bottomrule

Eye ID   &   $.754 \pm .005$ & $.778 \pm .003$ & $.789 \pm .001$ & $.795 \pm .001$                  \\ \hline
CST     &   $.694 \pm .004$ & $.721 \pm .003$ & $.739 \pm .001$ & $.749 \pm .001$                    \\
BCVA    &   $.760 \pm .009$ & $.788 \pm .001$ & $.783 \pm .001$ & $.790 \pm .001$                \\

BCVA + Eye ID  &   $.761 \pm .004$ & $.786 \pm .004$ & $.794 \pm .002$ & $.795 \pm .002$            \\

BCVA + CST  &   $.712 \pm .005$ & $.751 \pm .007$ & $.773 \pm .006$ & $.782 \pm .001$               \\
CST + Eye ID  &   $\textbf{.766} \pm .013$ & $\textbf{.786} \pm .003$ & $\textbf{.803} \pm .004$ & $\textbf{.806} \pm .003$             \\

BCVA + CST + Eye ID  &   $.747 \pm .005$ & $.778 \pm .003$ & $.802 \pm .004$ & $.806 \pm .002$        \\

 \bottomrule
\end{tabular}}
\caption{This table shows the average AUORC in a multi-label classification with different amounts of access to biomarker data for fine-tuning the model.}
\label{tab:biomarker_split_table}
\end{table}

Throughout the paper so far, all experiments have been done on a dataset that was based on the total \texttt{OLIVES} dataset that is derived from two clinical trials. However, in order to study the impact of reducing the amount of avilable data for contrastive pre-training, we extract the clinical labels and image from just the PRIME \cite{hannah2021real} clinical trial. this reduces the overal contrastive training pool to 29000 images. There are several interesting analytical points that we can draw from doing this. The first is that the PRIME subset has a wider variety of clinical information that we can potentailly use as a label. In addition to BCVA, CST, and Eye ID that are available across the whole \texttt{OLIVES} dataset; there are clinical parameters that exist for the PRIME subset specifically such as the type of diabetes of the patient, the diabetic retinopathy severity score (DRSS), and various demographic information. It can be observed that while CST, BCVA, and Eye ID still perform consistently well, there are other modalities that perform better than the self-supervised baselines. This includes leakage index and the number of years with diabetes. This indicates that there are potentially many unexplored clinical values in a variety of settings that have the potential to choose good positives and negatives for a contrastive loss. A possible reason for the huge difference in performance between the clinical and self-supervised methods is that Prime alone has fewer images as well as less image diversity compared to the total OLIVES dataset. As a result, regular self-supervised methods are more constrained in the representations that can be drawn from just taking augmentations to form positive pairs in a contrastive loss. In this way, they are more dependent on the total number of images in the training set as well as the total available data diversity.

We also analyze the impact of training with a progressively limited set of biomarker data.  To do this, we take our original training set of 7500 labeled biomarker scans and remove different sized subsets. This is shown in Table \ref{tab:biomarker_split_table} where each column represents the percentage amount of biomarker training data we restricted each method to have access to. We observe that using our contrastive learning strategy leads to improved performance in the multi-label classification setting. Additionally, strategies that make use of a combined clinical loss  again more consistently have a high performance even with reduced access to data for biomarker fine-tuning. Furthermore, for this table, we include a comparison with a network trained from scratch on the biomarker data only, using the same hyper-parameters as all previous experiments. This acts as an analysis of a semi-supervised setting since labeled data is progressively removed. It is interesting to note that the supervised methods, that have access to the biomarker labels during the entirety of training, are significantly worse as the training set is reduced. This shows the dependence that these methods have on a large enough training set because they are unable to leverage representations that may be learnt from the large unlabeled pool of data. The self-supervised methods we compare against are better able to make use of these representations to perform better on the smaller amount of available training data, but are still inferior to our method that integrates clinical labels into the contrastive learning process. An additional reason for this difference is the difficulty of the multi-label classification task. Within this task, certain biomarkers are easier to learn than others and it is possible that without sufficient training data these models more readily learn the features of the easier to detect classes and neglect the classes that exist as higher granular features. 

\begin{table}[]
\centering
\begin{tabular}{@{}ccc@{}}
\toprule
\multicolumn{3}{c}{Semi-Supervised Setting Performance (Accuracy/F1-Score)} \\ \midrule
                  & SimCLR v2 \cite{chen2020big}                 & CST + Eye ID               \\ \midrule
DME               & 68.66\% / .594             & \textbf{71.13}\% / \textbf{.696}             \\
IRF               & 63.73\% / .586             & \textbf{65.73}\% / \textbf{.590}             \\
IRHRF             & 57.13\% / .350             & \textbf{57.8}\% / \textbf{.553}              \\
FAVF              & 52.80\% / .503             & \textbf{54.30}\% / \textbf{.643}             \\
PAVF              & 47.60\% / \textbf{.183}             & \textbf{48.20}\% / .166             \\ \bottomrule
\end{tabular}
\caption{This table analyzes the integration of a clinical contrastive method into a state of the art semi-supervised framework \cite{chen2020big}.}
\label{tab:semi}
\end{table}

\subsection{Comparison with Eye Identity}
In \cite{vu2021medaug,diamant2021patient,azizi2021big}, the authors created a contrastive learning strategy based on choosing positives and negatives based on the patient identity. In order to create a comparison with these previously proposed approaches, we designate the strategy that chooses positives and negatives based on just the eye identity as its own separate section within each table that reflects the suggestion of these previous works. It can be observed that this method does perform better than the self-supervised baselines, as pointed out in previous work, but it does not do as well as our methods that make use of multiple clinical distributions such as CST + Eye ID. In Table \ref{tab:individual_biomarker_table}, methods that use other types of clinical labels or multiple clinical labels in tandem out-perform Eye ID on 4 out of the 5 biomarkers in terms of accuracy and f1-score. The same trend holds when looking at the patient split experiments in Table \ref{tab:patient_split_table}. Additionally, Eye ID consistently performs worse than methods that make use of multiple clinical distributions within the other robustness experiments that vary access to biomarker and clinical training data in Tables \ref{tab:biomarker_split_table} and \ref{tab:clinical_diverse_table}. This indicates that while patient/eye identity is an improvement over standard self-supervision, other individual clinical labels and their combinations can potentially offer better distributions on which positives and negatives for a contrastive loss can be sampled.

\subsection{Semi-Supervised Experiments}
We also compare our method within a state of the art semi-supervised framework in Table \ref{tab:semi}. In this case, we follow the setting of \cite{chen2020big}. To do this we take an encoder pre-trained with a contrastive learning strategy and fine-tune it with a linear layer with only 25\% of available biomarker data for each biomarker in this study. This model then becomes the teacher model that we use to train a corresponding student model. The student model has access to both the 25\% subset the teacher was trained on as well as the remaining biomarker data that we designate to be the unlabeled subset. The teacher is used to provide logit outputs that is then used as part of a distillation loss discussed in \cite{chen2020big} to train the student model. In this way, we model the semi-supervision setting by making use of a small amount of labeled data for the teacher model and then both labeled and unlabeled data for the student. We compare the performance of this setup on a pretraining with respect to SimCLR and our combined clinical contrastive strategy that makes use of both the CST and Eye ID label distributions. We observe in Table \ref{tab:semi}  that our method consistenly out-performs the model that used SimCLR pre-training. It is also interesting to note that the overall performance is lower than in the standard contrastive learning setting. Part of the reason for this is that this setup assumed access to a large unlabeled pool on the order of the size of ImageNet \cite{deng2009imagenet}. In this case, the teacher network may not have enough labels to initially fine-tune with and the corresponding distillation loss for the student network does not have enough unlabeled data to effectively impute the knowledge onto the student network. Despite the limitations of this constrained setting, we still see that our method is able to out-perform the SimCLR v2 baseline.

\subsection{Limitations}
While our analysis provides an in-depth look into the relationship between clinical and biomarker data for contrastive learning, there are still areas where further exploration is difficult to perform due to constraints of the medical setting. Specifically, part of the novelty of our analysis is derived from demonstrating the performance degradation of biomarkers as we approach higher levels of granularity. This shows that the conclusions of \cite{cole2021does} do have the potential to transfer within a medical setting. However, due to the difficulty of access to a sufficient amount of biomarker data we cannot perform an all-encompassing experiment of this granularity concept within a medical setting. Additionally, it is difficult to quantify this concept of granularity. We can intuitively get a sense of which biomarkers exhibit high and low granularity from medical studies such as \cite{markan2020novel}, but it remains somewhat hypothetical as to exact meaning of granularity. Furthermore, ideally this entire setting could be studied in many other contexts, but in most cases access to well-distributed clinical data as well as biomarkers is difficult to find in publicly available datasets which highlights the importance of the \texttt{OLIVES} dataset for this study. We encourage medical and machine learning practitioners to use the ideas presented in this paper as inspiration for the proper usage of clinical data within their own application settings.  

\section{Conclusion}

In this work, we investigate how the usage of a supervised contrastive loss on clinical data can be used to effectively train a model for the task of biomarker classification. We show how the method performs across different combinations of clinical labels, different architectures, different data access settings, and across different splits of patients or eye identities. We conclude that the usage of the clinical labels is a more effective way to leverage the correlations that exist within unlabeled data over traditional supervised and self-supervised algorithms, especially methods that make use of multiple clinical labels in a combined loss. We prove this through extensive experimentation on biomarkers of varying granularity within OCT scans and through this show that the granularity problem of contrastive learning exists within the medical domain as well. From a medical perspective, our paper shows that there are ways to utilize correlations that exist between measured clinical labels and their associated biomarker structures within images. Additionally, our method is based on practically relevant considerations regarding detecting key indicators of disease as well as challenges associated with labeling images for all the different manifestations of biomarkers that could be present. We hope this work inspires medical research into other domains and clinical settings where questions exist as to how to effectively utilize relationships that exist with the available data. 

\newpage

\bibliographystyle{IEEEbib}
\bibliography{ref}

\newpage
\begin{IEEEbiography}[{\includegraphics[width=1in,height=1in,clip,keepaspectratio]{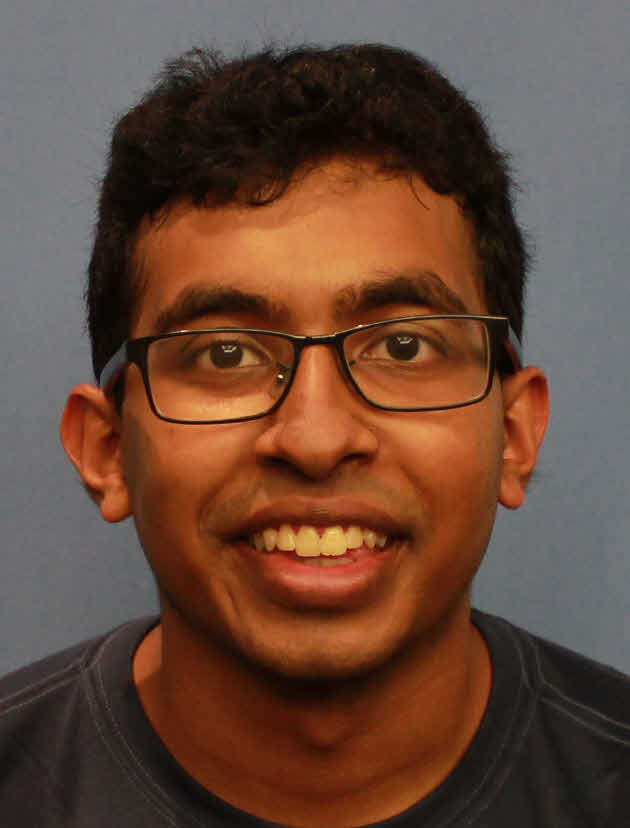}}]{Kiran Kokilepersaud}{\space} is a Ph.D. student in electrical and computer engineering at the Georgia Institute of Technology (Georgia Tech), Atlanta, Georgia, 30332, USA. He is currently a Graduate Research Assistant in the School of Electrical and Computer Engineering at the Georgia Institute of Technology in the Omni Lab for Intelligent Visual Engineering and Science (OLIVES) lab. He is a recipient of the Georgia Tech President's Fellowship for excellence amongst incoming Ph.D. students. His research interests include digital signal and image processing, machine learning, and its associated applications within the medical field.
\end{IEEEbiography}

\begin{IEEEbiography}[{\includegraphics[width=1in,height=1in,clip,keepaspectratio]{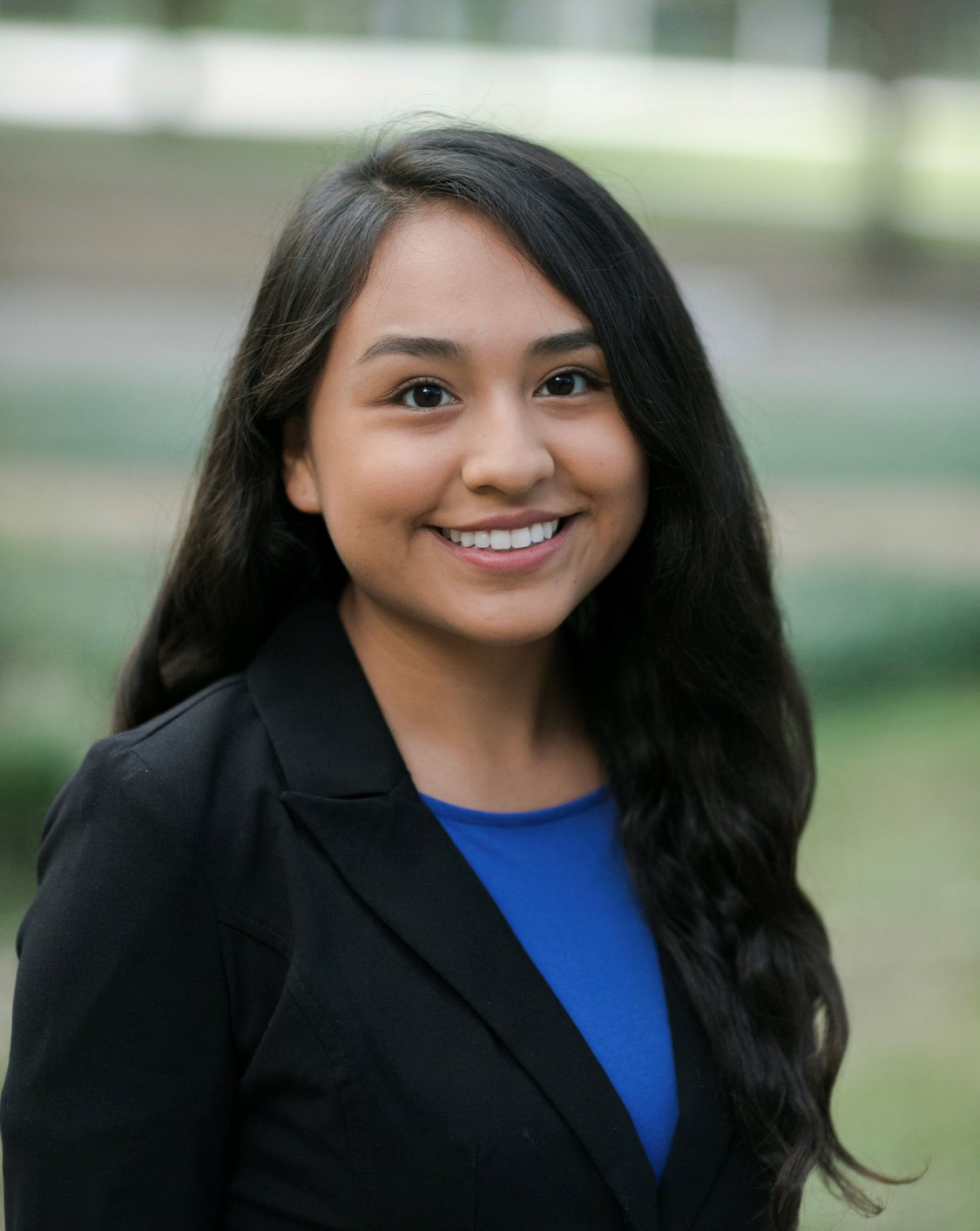}}]{Stephanie Trejo Corona}{\space} is a clinical research fellow at Retina Consultants of Texas. She graduated from Rice University with a B.S. in Biochemistry \& Cell Biology and a B.A. in Kinesiology in 2021. She has previously worked in the fields of synthetic and plant biology, with a passion for translational research. She was awarded the Rice University Biosciences Department Distinction in Research and Creative Work as well as multiple recognitions for her research presentations during her undergraduate career. She is currently working in the field of ophthalmology and hopes to improve healthcare access and clinical outcomes through prospective clinical trials, retrospective cohort studies, and applications of artificial intelligence.
\end{IEEEbiography}

\begin{IEEEbiography}[{\includegraphics[width=1in,height=1in,clip,keepaspectratio]{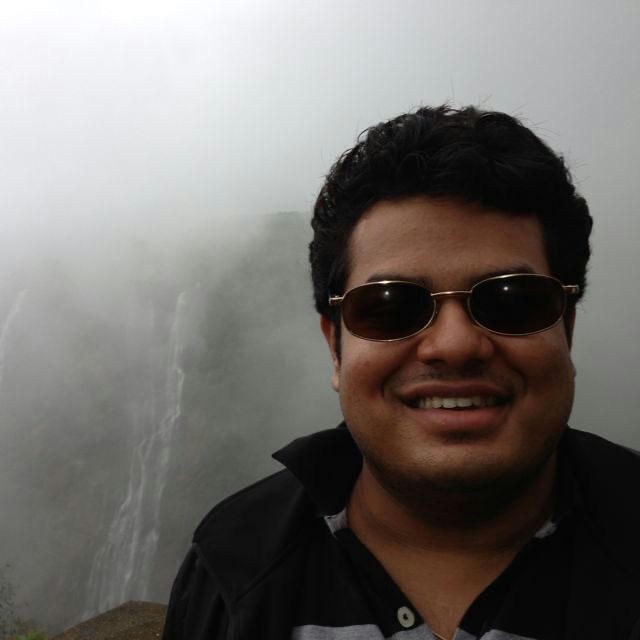}}]{Mohit Prabhushankar}{\space} received his Ph.D. degree in electrical engineering from the Georgia Institute of Technology (Georgia Tech), Atlanta, Georgia,
30332, USA, in 2021. He is currently a Postdoctoral Researcher and Teaching Fellow in the School of Electrical and Computer Engineering at the Georgia Institute of Technology in the Omni Lab for Intelligent Visual Engineering and Science (OLIVES) lab. He is working in the fields of image processing, machine learning, explainable and robust AI, active learning, and healthcare. He is the recipient of the Best Paper award at ICIP 2019 and Top Viewed Special Session Paper Award at ICIP 2020. He is the winner of the Roger P Webb ECE Graduate Research Excellence award in 2022. He is an IEEE Member.
\end{IEEEbiography}

\begin{IEEEbiography}[{\includegraphics[width=1in,height=1in,clip,keepaspectratio]{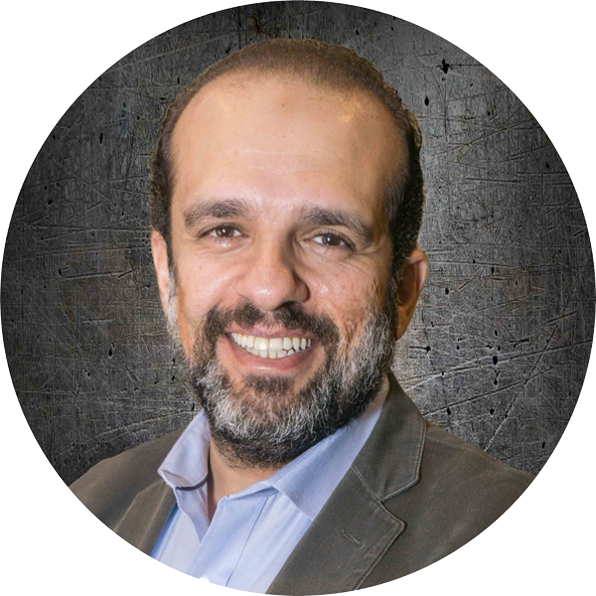}}]{Ghassan AlRegib}{\space} is the John and Marilu McCarty Chair Professor in the School of Electrical and Computer Engineering at the Georgia Institute of Technology. He was the recipient of the ECE Outstanding Junior Faculty Member Award, in 2008, and the 2017 Denning Faculty Award for Global Engagement. His research group, the Omni Lab for Intelligent Visual Engineering and Science (OLIVES) works on research projects related to explainable machine learning, robustness in intelligent systems, interpretation of subsurface volumes, and expanding healthcare access and quality. He has participated in several service activities within the IEEE and served on the editorial boards of several journal publications. He served as the TP co-Chair for ICIP 2020 and GlobalSIP 2014. He served as expert witness on several patents infringement cases and advised several corporations on both technical and educational matters. He is anIEEE Fellow.  
\end{IEEEbiography}

\begin{IEEEbiography}[{\includegraphics[width=1in,height=1in,clip,keepaspectratio]{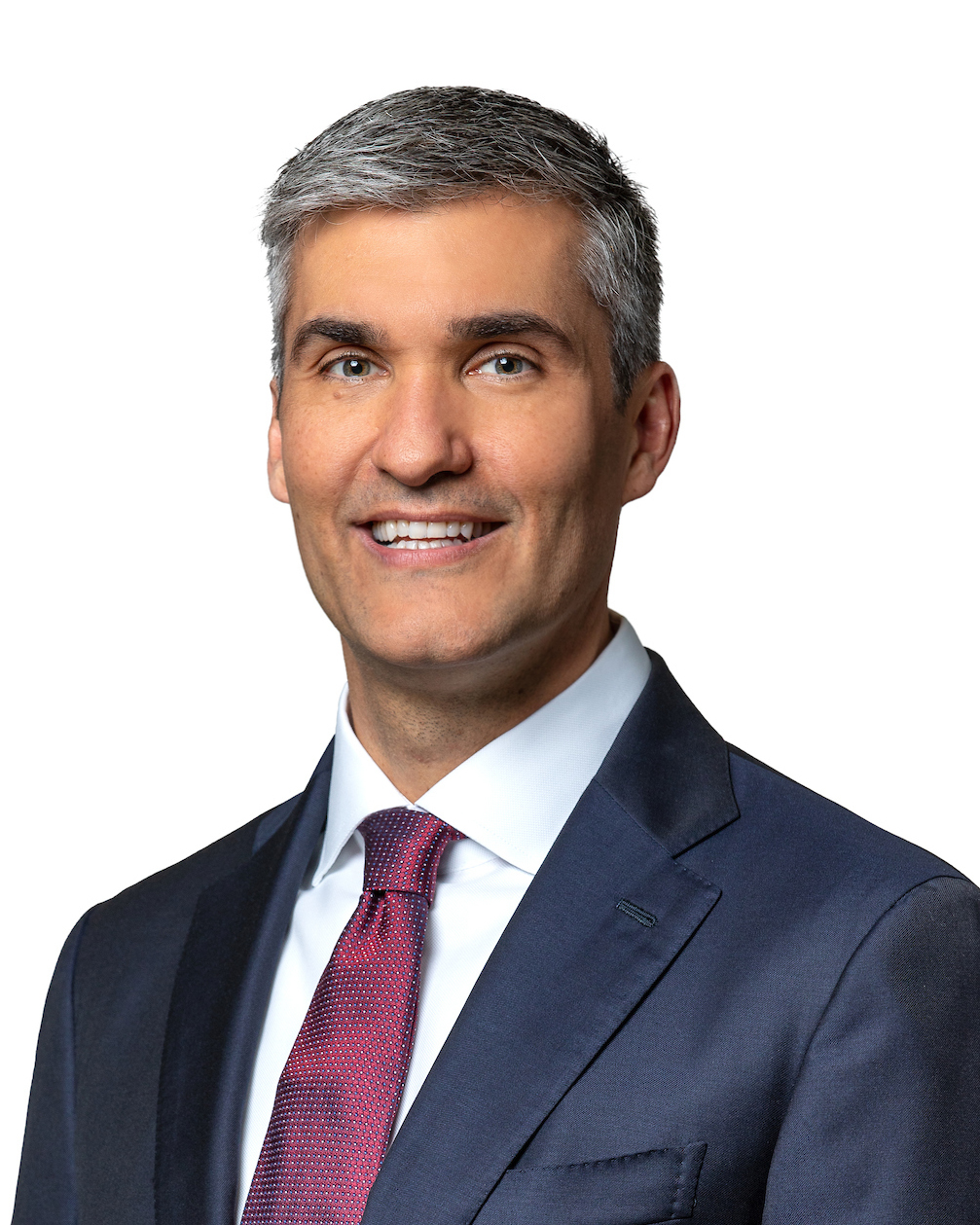}}]{Charles Wykoff}{\space},MD,PhD, is Director of Research at Retina Consultants of Texas; Chairman of Research, Retina Consultants of America; and Deputy Chair of Ophthalmology for the Blanton Eye Institute, Houston Methodist Hospital. He received his baccalaureate from MIT, PhD from Oxford, MD from Harvard, completed clinical training at Bascom Palmer Eye Institute, and is an active medical and surgical retina specialist. He is passionate about translational research, clinical trial design, accelerating drug-development programs and has published over 200 peer-reviewed manuscripts. He serves on multiple scientific and medical advisory boards, safety monitoring committees, and global steering committees for endeavors spanning the innovative process from early to late-stage developments. He is President of the Vit-Buckle Society (2021-2023), serves on the ASRS Board of Directors, is a founding member of the Ophthalmology Retina Editorial Board, and is the Chief Medical Editor for Retina Specialist. He has been awarded multiple Achievement, Honor and Senior Honor Awards including the ASRS Young Investigator and the AAO Secretariat Awards. His guiding philosophy is to build and strengthen innovative, ethical teams focused on developing new approaches to improving outcomes for blinding disease
\end{IEEEbiography}

\end{document}